\documentclass[conference]{IEEEtran}
\IEEEoverridecommandlockouts
\usepackage{cite}
\usepackage{amsmath,amssymb,amsfonts}
\usepackage{graphicx}
\usepackage{textcomp}
\usepackage{xcolor}

\usepackage{stfloats}
\usepackage{url}
\usepackage{verbatim}

\usepackage{cite}
\usepackage{amssymb}
\usepackage{booktabs}
\usepackage{adjustbox}

\usepackage{bbm}
\usepackage{arydshln}

 \usepackage{algpseudocode}
\usepackage[ruled]{algorithm}
\usepackage{subcaption}
\usepackage[rightcaption]{sidecap}
\usepackage{caption}
\usepackage{multirow}
\usepackage{hhline}
\usepackage{epstopdf}
\usepackage{tabularx}
\usepackage{tikz}
\usetikzlibrary{matrix}
\usepackage{mathtools, nccmath}
\usepackage{caption} 

\graphicspath{ {images/} }

\def\BibTeX{{\rm B\kern-.05em{\sc i\kern-.025em b}\kern-.08em
    T\kern-.1667em\lower.7ex\hbox{E}\kern-.125emX}}
\begin{document}

\title{Deep Learning on Hester Davis Scores for\\ Inpatient Fall Prediction}

\author{\parbox{16cm}{\centering
    {\large Hojjat Salehinejad$^{1,2}$,~\IEEEmembership{Senior Member,~IEEE}, Ricky Rojas$^{1}$, Kingsley Iheasirim$^{3}$, \\Mohammed Yousufuddin$^{4}$, and Bijan Borah$^{1}$}\\
    {\normalsize
    $^1$Kern Center for the Science of Health Care Delivery, Mayo Clinic, Rochester, MN, USA \\
    $^2$Department of Artificial Intelligence and Informatics, Mayo Clinic, Rochester, MN, USA \\    
    $^3$Department of Medicine, Mayo Clinic Health System, Mankato, MN, USA\\
    $^4$Department of Hospital Internal Medicine, Mayo Clinic Health System, Austin, MN, USA\\    
    \textit{\{salehinejad.hojjat, rojas.ricardo, iheasirim.kingsley,  yousufuddin.mohammed, borah.bijan\}@mayo.edu}
    }}
}

% \and
% \and
% \and
% \and
% \and

\maketitle

\begin{abstract}
Fall risk prediction among hospitalized patients is a critical aspect of patient safety in clinical settings, and accurate models can help prevent adverse events. The Hester Davis Score (HDS) is commonly used to assess fall risk, with current clinical practice relying on a threshold-based approach. In this method, a patient is classified as high-risk when their HDS exceeds a predefined threshold. However, this approach may fail to capture dynamic patterns in fall risk over time. In this study, we model the threshold-based approach and propose two machine learning approaches for enhanced fall prediction: One-step ahead fall prediction and sequence-to-point fall prediction. The one-step ahead model uses the HDS at the current timestamp to predict the risk at the next timestamp, while the sequence-to-point model leverages all preceding HDS values to predict fall risk using deep learning. We compare these approaches to assess their accuracy in fall risk prediction, demonstrating that deep learning can outperform the traditional threshold-based method by capturing temporal patterns and improving prediction reliability. These findings highlight the potential for data-driven approaches to enhance patient safety through more reliable fall prevention strategies.
\end{abstract}

\begin{IEEEkeywords}
Fall risk, fall prediction, Hester Davis score, machine learning. 
\end{IEEEkeywords}

\section{Introduction}

Fall risk assessment is a critical process in healthcare, aimed at identifying hospitalized patients who are at higher risk of falling during their stay~\cite{perell2001fall}. This assessment typically involves evaluating a combination of factors such as age, mobility, mental status, use of certain medications, and previous fall history. Tools like the Hester Davis Score (HDS) are widely employed to quantify fall risk based on these factors, allowing healthcare providers to classify patients into low, moderate, or high-risk categories~\cite{nyakundi2022use,payne2020impact}. By accurately identifying high-risk patients, hospitals can implement preventive measures, such as increasing monitoring, modifying the patient’s environment, or providing assistive devices to prevent falls. Effective fall risk assessment is key to improving patient safety, reducing fall-related injuries, and minimizing healthcare costs associated with prolonged hospital stays and related complications~\cite{perell2001fall,strini2021fall}.

The HDS is a widely used tool in clinical settings for fall risk evaluation, offering a structured and standardized scoring system that assesses key factors such as age, mental status, mobility, medication usage, continence, recent fall history, and behavioral tendencies. Each factor is assigned a weighted score based on its contribution to fall risk, producing a cumulative score that categorizes patients into risk levels. The HDS allows for near real-time reassessment, as it incorporates both static and dynamic characteristics of the patient. This facilitates timely interventions, such as bed alarms or increased supervision, to mitigate fall risks in hospitalized patients~\cite{hester2013validation}.

Despite the utility of the HDS and other threshold-based models, these approaches often fail to capture the evolving risk patterns over time. They rely on instantaneous values to trigger preventive measures, which may not reflect the subtle, progressive changes in a patient's condition. To address this limitation, machine learning models can offer a more dynamic and data-driven approach to fall risk prediction by incorporating the sequential pattern in the data. Similar approaches have been proposed for other challenges in healthcare, such as early warning systems~\cite{SALEHINEJAD2023102312,10363391}, hypertension detection~\cite{10313498}, and human activity recognition~\cite{9746803,10581974}. 

Machine learning, and particularly deep learning, has demonstrated superior performance in a variety of healthcare applications, from COVID-19 lung prognosis detection using chest computed tomography (CT) scans~\cite{lee2021deep}, to cervical spine fracture detection~\cite{salehinejad2021deep}, and in-hospital mortality prediction among diabetic intensive care unit (ICU) patients~\cite{theis2021improving}. In fall prediction, these models can be used to analyze complex, non-linear interactions between clinical variables, offering enhanced predictive power. 

In this paper, we model the traditional threshold-based fall risk assessment approach using HDS and propose two machine learning-based alternatives: One-step ahead fall prediction and sequence-to-point fall prediction using deep learning. The former uses the HDS at a given time to predict fall risk at the next timestamp, while the latter leverages all preceding samples in a time series to forecast fall events. Sequence-to-point prediction is particularly important, as it captures the entire sequence of events leading up to a fall, allowing the model to identify temporal patterns and trends that threshold-based methods might overlook. For example, a gradual increase in HDS values over time may signify rising fall risk, even if the individual scores do not exceed predefined thresholds. This approach enables more accurate and timely predictions, enhancing the ability to intervene before a fall occurs. Particularly, recurrent neural networks (RNNs)~\cite{haykin1995recurrent,salehinejad2017recent}, long short-term memory (LSTM)~\cite{sundermeyer2012lstm} networks, and gated recurrent unit (GRU)~\cite{dey2017gate} networks are proposed to learn from temporal pattern in the HDSs. We compare the performance of these methods to evaluate their potential in improving the accuracy and timeliness of fall risk predictions in clinical settings.
 
The source code used in this project and data can be made available on reasonable request and approval of corresponding authorities by contacting the corresponding author.

\begin{figure*}[t!]
    \centering
    \begin{subfigure}{0.7\textwidth}
        \includegraphics[width=\linewidth, trim={0.0cm 0.0cm 0.0cm 0.0cm}, clip]{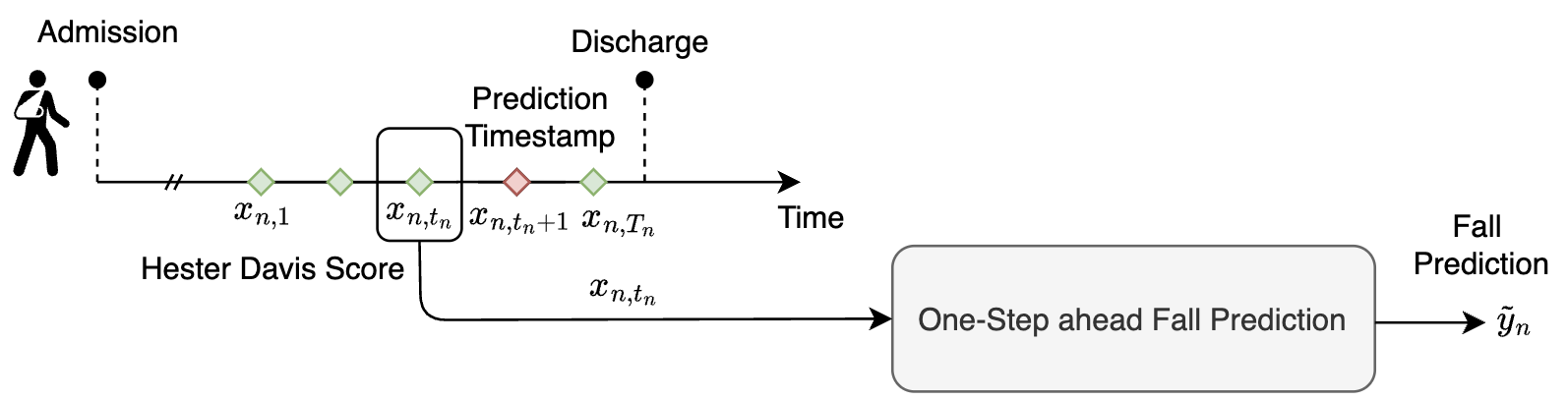}
        \caption{One-step ahead fall prediction model.}
        \label{fig:onestep}
    \end{subfigure}

    \begin{subfigure}{0.7\textwidth}
        \includegraphics[width=\linewidth, trim={0.50cm 0.0cm 0.0cm 0.0cm}, clip]{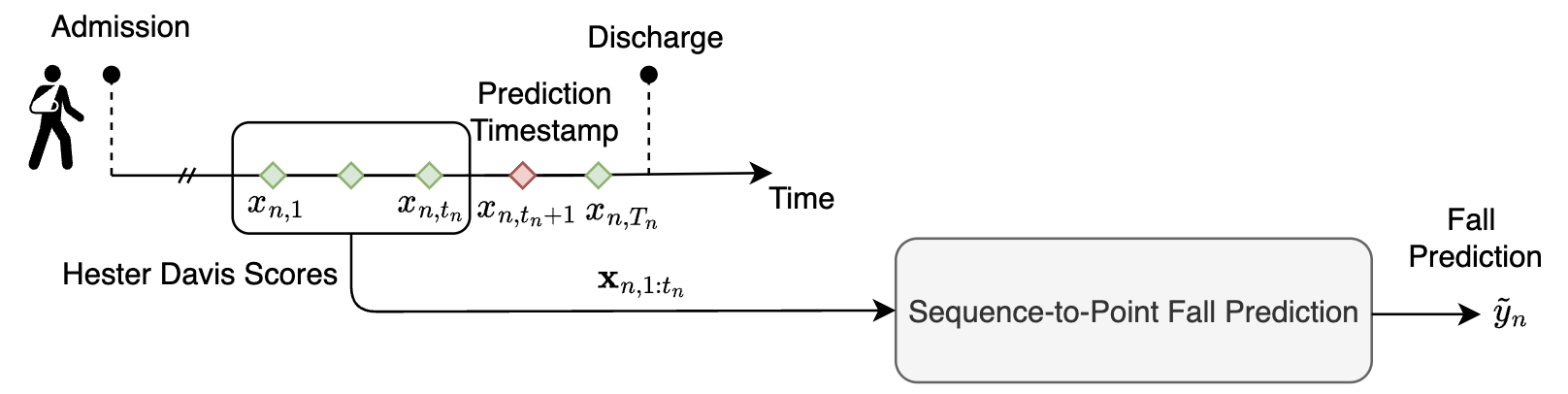}
        \caption{Sequence-to-point fall prediction model.}
        \label{fig:sequence}
    \end{subfigure}
    \caption{Proposed schemes for learning from Hester Davis scores (HDSs) in one-step ahead and sequence-to-point modes. The objective is to predict fall event at the prediction timestamp $x_{n,t_{n+1}}$ using prior HDSs.}
    \label{fig:fallModel}
\end{figure*}

\section{Fall Prediction Models}

In this section, fall prediction using HDS is modeled in two schemes, as illustrated in Figure~\ref{fig:fallModel}. Let ${\mathbf{x}_n=(x_{n,1},...,x_{n,t},...,x_{n,T})}$ represent $T$ HDSs of an individual $n\in\{1,...,N\}$ from admission to discharge, where $t+1$ is a prediction timestamp, $T$ is the total number of retrospective samples, and $N$ is the total number of individuals. The HDSs of patients are calculated every $\Delta T$ hours. 
In the One-step ahead fall prediction scheme, in order to make a prediction for an individual $n$ at the future timestamp $t_n+1$, the last HDS $x_{n,t_n}$ is used.  In the sequence-to-point fall prediction scheme, the entire HDS samples $(x_{n,1},...,x_{n,t_n})$ since admission, are used to make a prediction for an individual $n$ at the future timestamp $t_n+1$.

\subsection{One-Step ahead Fall Prediction}
The current clinical practice approach involves comparing the HDS to a predefined threshold. In this section, we mathematically model this approach and then propose machine learning models to learn from the HDS at the current timestamp $t$ in order to predict the outcome at the subsequent timestamp $t+1$.

\subsubsection{Threshold-based Method}
\label{sec:rule_based}
Most clinical providers use an absolute number threshold $\theta$ to determine if a patient is at a high risk of fall and needs extra care and increased monitoring. In this approach, if at any time $t$ the HDS value  $x_{n,t}$ for patient $n$ exceeds the threshold $\theta$, the patient is classified as high-risk for falls, defined as
\begin{equation}
\Tilde{y}_{n,t+1}= \left\{ 
  \begin{array}{ c l }
    1 & \quad \textrm{if } x_{n,t_n} \geq \theta \\
    0                 & \quad \textrm{otherwise}
  \end{array},
\right.   
\end{equation}
where $\Tilde{y}_{n,t+1}=1$ means the patient $n$ is at high-risk of fall at the future timestamp $t_n+1$ and $\Tilde{y}_{n,t+1}=0$ means otherwise.

\subsubsection{Machine Learning Methods}
It is possible to build a binary fall prediction model about the fall outcome at a future timestamp $t+1$ based solely on the current available HDS sample $x_{n,t}$. The task is to predict a binary label ${\tilde{y}_{n,t+1}\in \{0, 1\}}$ at time $t+1$, using only the value of $x_{n,t}$, the sample immediately preceding $t+1$. For each sample $x_{n,t}$ in the retrospective dataset, we pair it with a corresponding label $y_{n,t+1}$, which represents the outcome at the next time step. The prediction model is built using a binary classifier $\phi(x_{n,t})$, which maps each $x_{n,t}$ to a binary outcome $\Tilde{y}_{n,t+1}$ as
\begin{equation}
\Tilde{y}_{n,t+1} = \phi(x_{n,t}),
\end{equation}
where the training set consists of $N-1$ pairs $\{(x_{1,t}, y_{1,t+1}), \dots, (x_{N-1,t}, y_{N-1,t+1})\}$ and each $x_{n,t}$ serves as a feature to predict the binary outcome $y_{n,t+1}$. The classifier is trained to minimize the prediction error by adjusting its parameters to best capture the relationship between the single time series sample $x_{n,t}$ and the next time step's binary fall label. Various machine learning models such as k-nearest neighbors (KNN)~\cite{laaksonen1996classification}, support vector machine (SVM)~\cite{smola2004tutorial}, random forest (RF)~\cite{breiman2001random}, and extreme gradient boosting (XGB)~\cite{chen2016xgboost} are evaluated in experiments section for this aim.

\subsection{Sequence-to-Point Fall Prediction}
Sequence-to-point fall prediction, which utilizes all preceding samples in a time series to predict a fall event, can hold significant importance in clinical settings. Unlike traditional threshold-based methods that rely on single, instantaneous values, sequence-to-point prediction leverages the entire sequence of data leading up to the instant before which the fall event is predicted. This approach captures temporal patterns and trends that may be missed using isolated samples.

\subsubsection{Recurrent Neural Networks} 
The RNNs are one of the popular methods to model sequential dependencies within time series, making it suitable for tasks where the prediction at the final time step depends on the prior inputs. These networks leverages its hidden state to capture the temporal dynamics from sequential inputs, enabling the model to predict the risk of a fall at the final time step.

In order to model the sequence-to-point binary classification task for inpatient fall risk prediction using RNNs~\cite{salehinejad2017recent}, let $x_t$ represent a HDS at time $t$ for an individual without loss of generality. The RNN processes each time series up to time step $t$ to predict whether a fall occurs at time $t+1$. The hidden state $\mathbf{h}_t$ at each time step is computed as
\begin{equation}
 \mathbf{h}_t = \sigma(\mathbf{W}_{IH} \cdot \textbf{x}_t + \mathbf{W}_{HH} \cdot \mathbf{h}_{t-1} + \mathbf{b}_h),   
\end{equation}
 where $\sigma(\cdot)$ is the activation function, $\mathbf{h}_t$ is the hidden state at time step $t$, $\mathbf{W}_{IH}$ is the input weight matrix, $\mathbf{W}_{HH}$ is the recurrent weight matrix, and $\mathbf{b}_h$ is the bias vector. The hidden state $\mathbf{h}_{t}$ at time step $t$  captures the information of the HDSs up to that point. 

At the time step $t$, the hidden state $\mathbf{h}_{t}$ is used to predict the occurrence of a fall at time step $t+1$. The hidden state is passed through a fully connected layer followed by a Softmax activation to produce the output logits  as
\begin{equation}
    \mathbf{z}_{t+1} = \mathbf{W}_{HO} \cdot \mathbf{h}_{t} + \mathbf{b}_o,
    \label{eq:z}
\end{equation}
where $\mathbf{W}_{HO}$ is the output weight matrix and $\mathbf{b}_o$ is the bias term. The Softmax activation function is applied to the logits to obtain the probability distribution over the two classes (fall or no fall) as
\begin{equation}
p_c = \frac{e^{z_{t+1, c}}}{\sum_{c=1}^2 e^{z_{t+1, c}}},
\end{equation}
where $z_{t+1, c}$ is the logit corresponding to class $c$ (fall or no fall)and the predicted outcome is
\begin{equation}
    \tilde{y}_{n,t+1}=\arg\max_{c \in \{1, 2\}} p_c.
\end{equation}
For simplicity, assume $p$ as the predicted probability of the fall outcome class. The networks is trained using backpropagation and cross-entropy loss function as
\begin{equation}
   \mathcal{L} = - \frac{1}{N} \sum_{n=1}^{N} \left( y_n \cdot \log(p_n) + (1 - y_n) \cdot \log(1 - p_n) \right),
   \label{eq:loss}
\end{equation}
where $y_{n}\in\{0,1\}$ is the true label, $p_n$ is the predicted probability of the fall outcome class of individual $n\in\{1,...,N\}$.

\subsubsection{Long Short-Term Memory Networks}
To address the issue of vanishing gradients commonly faced by standard RNNs, we implemented a LSTM network, which introduces gates to control information flow and maintain long-range dependencies across time steps~\cite{salehinejad2017recent}. It maintains an internal memory state $\mathbf{c}_t$ along with the hidden state $\mathbf{h}_t$. At each time step $t$, the LSTM computes the input gate as
\begin{equation}
    \mathbf{g}_t^{i} = \sigma(\mathbf{W}_{Ig^{i}} \cdot \textbf{x}_t + \mathbf{W}_{Hg^{i}} \cdot \mathbf{h}_{t-1} + \mathbf{b}_{g^i}), 
\end{equation}
where $\mathbf{W}_{Ig^{i}}$ is the weight matrix from the input layer to the input gate, $\mathbf{W}_{Hg^{i}}$ is the weight matrix from hidden state to the input gate, and $\mathbf{b}_{g^i}$ is the bias of the input gate. The forget gate is defined as
 \begin{equation}
\mathbf{g}_t^{f} = \sigma(\mathbf{W}_{Ig^{f}} \cdot \textbf{x}_t + \mathbf{W}_{Hg^{f}} \cdot \mathbf{h}_{t-1} + \mathbf{b}_{g^f}), 
\end{equation}
where $\mathbf{W}_{Ig^{f}}$ is the weight matrix from the input layer to the
forget gate, $\mathbf{W}_{Hg^{f}}$ is the weight matrix from hidden state
to the forget gate, and $\mathbf{b}_{g^f}$ is the bias of the forget gate. The cell gate as
\begin{equation}
\mathbf{g}_t^{c} = \tanh(\mathbf{W}_{Ig^{c}} \cdot \textbf{x}_t + \mathbf{W}_{Hg^{c}} \cdot \mathbf{h}_{t-1} + \mathbf{b}_{g^c}), 
\end{equation}
 where $\mathbf{W}_{Ig^{c}}$ is the weight matrix from the input layer to the
cell gate, $\mathbf{W}_{Hg^{c}}$ is the weight matrix from hidden state to the
cell gate, and $\mathbf{b}_{g^c}$ is the bias of the cell gate. The output gate is 
\begin{equation}
\mathbf{g}_t^{o} = \sigma(\mathbf{W}_{Ig^{o}} \cdot \textbf{x}_t + \mathbf{W}_{Hg^{o}} \cdot \mathbf{h}_{t-1} + \mathbf{b}_{g^o}), 
\end{equation}
where $\mathbf{W}_{Ig^{o}}$ is the weight matrix from the input layer to the
output gate, $\mathbf{W}_{Hg^{o}}$ is the weight matrix from hidden state
to the output gate, and $\mathbf{b}_{g^o}$ is the bias of the output gate. 

The memory state $\mathbf{c}_t$ and hidden state $\mathbf{h}_t$ are updated as 
\begin{equation}
\mathbf{g}_t^{c}= \mathbf{g}_t^{f} \odot \mathbf{g}_{t-1}^{c}+ \mathbf{g}_t^{i}\odot \mathbf{g}_t^{\tilde{c}},
\end{equation}
and
\begin{equation}
\mathbf{h}_t = \mathbf{g}_t^{o} \odot \tanh(\mathbf{g}_t^{c}),
\end{equation}
where $\mathbf{g}_t^{\tilde{c}}$ is the candidate cell state~\cite{salehinejad2017recent}. At time step $t$, the hidden state $\mathbf{h}_t$ is used to predict the fall event similar to Eqs.~(\ref{eq:z}) and~(\ref{eq:loss}) in training the proposed RNN.

\subsubsection{Gated Recurrent Unit}

The GRU is a simplified variant of the LSTM that reduces the number of gates while retaining the ability to manage long-range dependencies~\cite{salehinejad2017recent}. GRUs simplify the gating mechanism by combining the forget and input gates into a single update gate.
At each time step $t$, the GRU computes an update gate as
\begin{equation}
    \mathbf{z}_t = \sigma(\mathbf{W}_z \cdot \mathbf{x}_t + \mathbf{U}_z \cdot \mathbf{h}_{t-1} + \mathbf{b}_z),
\end{equation}
and the reset gate as
\begin{equation}
\mathbf{r}_t = \sigma(\mathbf{W}_r \cdot \mathbf{x}_t + \mathbf{U}_r \cdot \mathbf{h}_{t-1} + \mathbf{b}_r), 
\end{equation}
and the candidate hidden state as 
\begin{equation}
\tilde{\mathbf{h}}_t = \tanh(\mathbf{W}_h \cdot \mathbf{x}_t + \mathbf{U}_h \cdot (\mathbf{r}_t \odot \mathbf{h}_{t-1}) + \mathbf{b}_h),
\end{equation}
where the hidden state is then updated as
\begin{equation}
\mathbf{h}_t = (1 - \mathbf{z}_t) \odot \mathbf{h}_{t-1} + \mathbf{z}_t \odot \tilde{\mathbf{h}}_t.
\end{equation}
At time step $t$, the hidden state $\mathbf{h}_t$ is used to predict the fall event at timestamp $t+1$ similar to Eqs.~(\ref{eq:z}) and~(\ref{eq:loss}) in training the RNN.

\section{Experiments}

\subsection{Data}
Our Institutional Review Board approved the study protocol. The dataset consisted of $46,695$ hospitalized patients, including $4,245$ who experienced a fall (median age $66$; $44\%$ male) and $42,450$ who did not (median age $66$; $48\%$ male).
Retrospective data was collected from consecutive patients admitted between January 1, 2018, and May 23, 2023, for various medical and surgical conditions across 4 academic and 13 community hospitals in Arizona, Florida, Minnesota, and Wisconsin in the United States. Patients were identified using electronic medical records. Adults aged 18 years and older who had been hospitalized for at least one day were included in the study, while those admitted to critical care units, hospice, or psychiatric units were excluded.

\begin{table*}[]
\captionsetup{font=footnotesize}
\caption{Performance results in \textbf{one-step ahead fall prediction}. Results are normalized to a scale of one and averaged over 10-fold cross-validation.}
\centering
\begin{adjustbox}{width=0.67\textwidth}
\begin{tabular}{ccccccc}
\toprule
 \multirow{2}{*}{Model} & \multicolumn{5}{c}{Performance Metric (Avg.$\pm$Std.)} \\ \cmidrule{2-7} 
     & \text{Accuracy} & \text{F1 Score} & \text{Specificity} & \text{Sensitivity} & \text{PPV} & \text{AUC} \\ \midrule
\text{HDS 7} & 0.57$\pm$0.01 & 0.57$\pm$0.01 & 0.52$\pm$0.01 & 0.62$\pm$0.01 & 0.54$\pm$0.01 & 0.57$\pm$0.01 \\ \midrule
\text{HDS 20} & 0.60$\pm$0.01 & 0.56$\pm$0.00 & 0.92$\pm$0.00 & 0.29$\pm$0.01  & 0.65$\pm$0.01 & 0.60$\pm$0.01\\ \midrule
\text{KNN} & 0.52$\pm$0.00 & 0.39$\pm$0.01 & 0.99$\pm$0.01 & 0.05$\pm$0.01  & 0.05$\pm$0.01& 0.54$\pm$0.01 \\ \midrule
\text{SVM} & 0.63$\pm$0.01 & 0.62$\pm$0.01 & 0.82$\pm$0.01 & 0.44$\pm$0.03 & 0.57$\pm$0.01& 0.66$\pm$0.01  \\ \midrule
\text{RF} & 0.63$\pm$0.01 & 0.62$\pm$0.01 & 0.80$\pm$0.01 & 0.46$\pm$0.01 & 0.62$\pm$0.01 & 0.70$\pm$0.01 \\ \midrule
\text{XGB} & 0.63$\pm$0.01 & 0.62$\pm$0.01 & 0.81$\pm$0.01 & 0.46$\pm$0.01  & 0.62$\pm$0.01& 0.70$\pm$0.01 \\ \bottomrule

\end{tabular}
\end{adjustbox}
\label{T:single_value_HDS_results}
\end{table*}

\begin{table*}[]
\captionsetup{font=footnotesize}
\caption{Performance results in \textbf{sequence-to-point fall prediction}. Results are normalized to a scale of one and averaged over 10-fold cross-validation.}
\centering
\begin{adjustbox}{width=0.67\textwidth}
\begin{tabular}{ccccccc}
\toprule
 \multirow{2}{*}{Model} & \multicolumn{5}{c}{Performance Metric (Avg.$\pm$Std.)} \\ \cmidrule{2-7} 
     & \text{Accuracy} & \text{F1 Score} & \text{Specificity} & \text{Sensitivity} & \text{PPV} & \text{AUC} \\ \midrule
        
RNN & 0.69$\pm$0.13 & 0.64$\pm$0.18 & 0.69$\pm$0.24 & 0.69$\pm$0.37 & 0.69$\pm$0.19 & 0.66$\pm$0.12 \\ \midrule
LSTM & 0.70$\pm$0.12 & 0.66$\pm$0.18 & 0.64$\pm$0.22 & 0.44$\pm$0.27 & 0.76$\pm$0.12 & 0.70$\pm$0.10 \\ \midrule
GRU & 0.74$\pm$0.20 & 0.67$\pm$0.28 & 0.94$\pm$0.07 & 0.53$\pm$0.44 & 0.53$\pm$0.18 & 0.77$\pm$0.09\\ \bottomrule
        
\end{tabular}
\end{adjustbox}
\label{T:ts_HDS_results}
\end{table*}

\subsection{Evaluation Setup}

A 10-fold cross-validation was performed, with the average (Avg.) and standard deviation (Std.) of each performance metric recorded. In each independent run, the models were trained from scratch on a randomly selected training dataset and evaluated on a randomly selected balanced test dataset. For each cross-validation fold, a balanced test set was created by randomly selecting 10\% of the data from the fall event class and 10\% from the no fall event class. This left the remaining dataset imbalanced. To address this, a balanced training dataset was constructed for each fold by including all remaining encounters from the fall event class and randomly selecting an equal number of encounters from the no fall event class. The combined samples were shuffled prior to each training iteration.

The machine learning models were evaluated using several metrics. Accuracy is defined as
\begin{equation}
Acc = \frac{TP+TN}{P+N},
\end{equation}
where $TP$ is the true positive value, $TN$ is the true negative value, $P$ is the number of true fall encounters, and $N$ is the number of true encounters without a fall. With a balanced test dataset, accuracy equals balanced accuracy. The F1 Score is given by 
\begin{equation}
F1 = \frac{2\cdot TP}{2\cdot TP+FP+FN},   
\end{equation}
where $FP$ represents false positives (encounters incorrectly predicted as fall event) and $FN$ denotes false negatives (encounters incorrectly predicted as not fall). Specificity, or true negative rate, is defined as 
\begin{equation}
TNR = \frac{TN}{TN+FP},   
\end{equation}
and sensitivity, or true positive rate, is calculated as 
\begin{equation}
TPR = \frac{TP}{TP+FN}.   
\end{equation}
The Positive Predictive Value (PPV) is defined as the proportion of $TP$s out of the total number of positive results, calculated as
\begin{equation}
PPV = \frac{TP}{TP + FP}.    
\end{equation}

\subsection{Training Setup}

The hyperparameter tuning was conducted using $10\%$ of the training data as the validation dataset, which was different from the test dataset, using random serach. All the models were implemented in Python and PyTorch~\cite{paszke2019pytorch} and trained on two NVIDIA A6000 GPUs with $256$GB of RAM and $64$ CPU cores. 

The SVM~\cite{hearst1998support} model was built with a radial basis function (RBF) with a regularization parameter of $0.1$ (grid searched in $\{0.001,0.01,0.1,1,10\}$). The KNN model was evaluated for various nearest neighbor 
values in $\{1, 2, ..., 10\}$ and the 1-nearest neighbour was selected. The XGB model was trained with $300$ estimators, $2$ parallel trees, and regularization coefficient $1$. The number of trees in RF was searched ranging from $100$ to $500$ with step $100$, set to $300$, and the maximum depth of trees was set to $10$ to prevent overfitting. 

 Hyperparameter tuning for RNNs, LSTMs, and GRUs involved selecting optimal values for each parameter to maximize model performance and efficiency. For the number of units, the search was in $\{32, 64, 128, 256\}$ with one layer.  Exponential adaptive learning rate with Adam optimizer was used initiated from $0.1$ and $0.01$, with $0.1$ being a good starting point, with a batch size of $32$. Early-stopping was applied with a patience of $5$ epoch given $200$ training epochs. Dropout~\cite{labach1904survey} rate was set to $0.5$. In LSTMs, setting the forget gate bias close to $1$ helped the model retain long-term dependencies, and in GRUs, adjusting the update gate similarly enhances performance. The rectified linear unit (ReLU)~\cite{karim2017lstm} activation function was used due to its non-linearity and faster convergence.

\subsection{Performance Results Analysis}
Table \ref{T:single_value_HDS_results} presents the performance results for various models in one-step-ahead fall prediction, with metrics normalized to a scale of 1 and averaged over 10-fold cross-validation. The models evaluated include two threshold-based methods, the HDS 7 and HDS 20, and several machine learning algorithms including KNN, SVM, RF, and XGB. The HDS 7 threshold-based method demonstrates moderate and consistent performance across all metrics, while HDS 20 achieves high specificity at $0.92$ but low sensitivity at $0.29$, indicating its strength in identifying non-fall events at the expense of detecting actual falls. KNN exhibits the highest specificity at $0.99$ but suffers from extremely low sensitivity at $0.05$, highlighting its poor performance in fall detection. In contrast, the machine learning models SVM, RF, and XGB show comparable performance, with accuracy around $0.63$, balanced F1 scores, and the area under the curve (AUC) values ranging from $0.66$ to $0.70$. These results indicate that these models are more effective in balancing sensitivity and specificity compared to the threshold-based methods, with RF and XGB providing the best overall discriminative power, as evidenced by their higher AUC scores at $0.70$.

Table \ref{T:ts_HDS_results} presents the performance results for various models in sequence-to-point fall prediction. Among the LSTM, GRU, and RNN models, distinct differences in effectiveness are observed. The GRU model achieves the highest accuracy at $0.74$, demonstrating its strong capability to classify instances correctly. It also shows a commendable F1 score of $0.67$, reflecting a balance between precision and recall, alongside impressive specificity at $0.94$ and moderate sensitivity at $0.53$. Conversely, the RNN model, with slightly lower accuracy at $0.69$, demonstrates higher sensitivity at $0.69$, suggesting better performance in identifying positive instances. The LSTM model exhibits competitive performance with an accuracy of $0.70$ and a favorable F1 score of $0.66$, though its specificity and sensitivity indicate a trade-off in accurately identifying true negatives and positives.

The superior performance of the GRU compared to the RNN and LSTM can be attributed to its streamlined architecture, which employs fewer parameters while effectively capturing long-range dependencies. By combining the forget and input gates into a single update gate, the GRU simplifies the model, enhancing its learning efficiency. This design helps mitigate the vanishing gradient problem that often plagues traditional RNNs, allowing the GRU to converge faster during training. Additionally, the GRU typically requires less computational resources, making it an attractive option in scenarios where model efficiency is critical.

\begin{figure}[!t]
\centering
\captionsetup{font=footnotesize}
\centering
\includegraphics[width=0.48\textwidth]{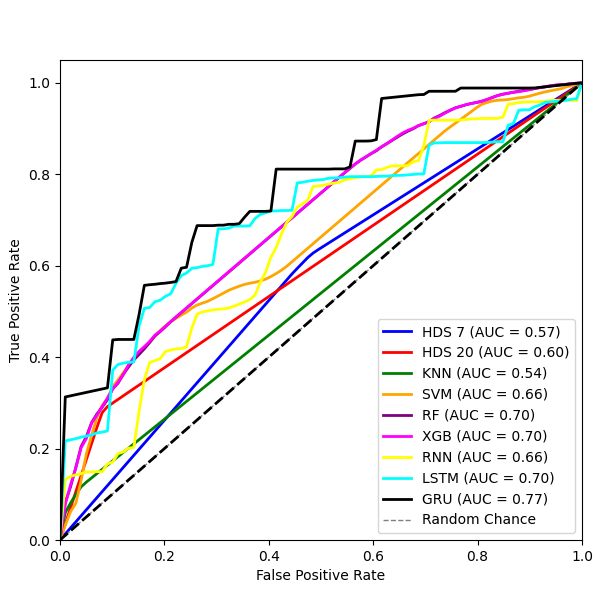}
\caption{Receiver operating characteristic curve
(ROC) of models.} 
\label{fig:roc}
\end{figure}

Figure~\ref{fig:roc} displays the receiver operating characteristic curve (ROC) curves of models. Overall, the GRU model stands out for its accuracy and specificity, making it a preferable choice for applications prioritizing precision. However, if maximizing the identification of positive cases is the primary goal, the RNN model may be more suitable due to its higher sensitivity. Therefore, the selection of the model should be guided by the specific requirements of the application, whether focusing on maximizing correct classifications or optimizing for sensitivity.

\section{Conclusion}
In conclusion, effective fall risk assessment is crucial for enhancing patient safety in healthcare settings, particularly for hospitalized individuals. Traditional methods, such as the threshold-based HDS, provide a structured approach to evaluating fall risk but often fall short in capturing the dynamic nature of patient conditions. This study highlights the limitations of threshold-based models, which may overlook subtle changes in risk factors over time. In contrast, machine learning approaches, including one-step ahead and sequence-to-point fall prediction methods, offer a more sophisticated framework for predicting fall risk by analyzing temporal patterns and interactions among clinical variables. The comparative analysis demonstrates that machine learning models, particularly the GRU, outperformed traditional methods and provided a more balanced sensitivity and specificity. By utilizing advanced algorithms, healthcare providers can achieve more accurate predictions, leading to timely interventions that can significantly reduce the incidence of falls and associated complications. Future research should continue to explore and refine these machine learning techniques to further enhance fall risk assessment strategies in clinical practice.

{\small
\bibliographystyle{unsrt}
\bibliography{main}
}

\end{document}